\pdfoutput=1

\documentclass[11pt]{article}

\usepackage[]{acl}

\usepackage{times}
\usepackage{latexsym}
\usepackage{algorithm}
\usepackage[noend]{algpseudocode}
\usepackage{algorithmicx,algorithm}

\usepackage[T1]{fontenc}

\usepackage[utf8]{inputenc}

\usepackage{microtype}
\usepackage{graphicx}
\usepackage{booktabs,fancyvrb}
\usepackage{multirow}
\usepackage{inconsolata}
\usepackage{xspace}
\usepackage{subcaption}
\usepackage{amsmath}
\usepackage{cleveref}
\usepackage{paralist}
\usepackage{enumitem}
\usepackage[normalem]{ulem}
\usepackage{xparse}
\usepackage{makecell}
\usepackage[figuresright]{rotating}
\title{Measuring and Improving Compositional Generalization in Text-to-SQL via Component Alignment}

\author{Yujian Gan${}^{1}$ \ \ \ \ Xinyun Chen${}^{2}$ \ \ \ \ Qiuping Huang${}^{4}$ \ \ \ \  Matthew Purver${}^{1,3}$ \\
${}^{1}$Queen Mary University of London \ \ \ \  \ \ \ \ ${}^{2}$UC Berkeley
\ \ \  \ \ \ \ ${}^{3}$Jožef Stefan Institute\\
${}^{4}$Nanning Central Sub-branch of the People's Bank of China \\
  \texttt{\{y.gan,m.purver\}@qmul.ac.uk}  \ \ \ \  \ \ \ \  \texttt{xinyun.chen@berkeley.edu} \\
 \texttt{qiuping\_h@foxmail.com} \ \  \ \  
}

\begin{document}
\maketitle
\begin{abstract}
In text-to-SQL tasks --- as in much of NLP --- \emph{compositional generalization} is a major challenge: neural networks struggle with compositional generalization where training and test distributions differ.
However, most recent attempts to improve this are based on word-level synthetic data or specific dataset splits to generate compositional biases.
In this work, we propose a clause-level compositional example generation method. 
We first split the sentences in the Spider text-to-SQL dataset into sub-sentences, annotating each sub-sentence with its corresponding SQL clause, resulting in a new dataset Spider-SS.
We then construct a further dataset, Spider-CG, by composing Spider-SS sub-sentences in different combinations, to test the ability of models to generalize compositionally.
Experiments show that existing models suffer significant performance degradation when evaluated on Spider-CG, even though every sub-sentence is seen during training.
To deal with this problem, we modify a number of state-of-the-art models to train on the segmented data of Spider-SS, and we show that this method improves the generalization performance.\footnote{Our code and dataset are available at  \href{https://github.com/ygan/SpiderSS-SpiderCG}{https://github.com/ygan/SpiderSS-SpiderCG}}
\end{abstract}

\section{Introduction}
\label{Section:intro}
Neural models in supervised learning settings show good performance on data drawn from the training distribution.
However, generalization performance can be poor on out-of-distribution (OOD) samples~\cite{Finegan-Dollak2018,Suhr2020,Kaushik2020Learning,sagawa2020distributionally}.
This might be the case even when the new samples are composed of known constituents; e.g., on the SCAN dataset \cite{pmlr-v80-lake18a}, many models give incorrect predictions for the input ``jump twice and walk'', even when ``jump twice'', ``walk'', and ``walk twice'' are seen during training. 
This (often lacking) ability to generalize to novel combinations of elements observed during training is referred to as \emph{compositional generalization}.
 

Previous work on compositional generalization in text-to-SQL focuses on query split.
For example, \citet{shaw-etal-2021-compositional} propose TMCD split based on SQL atoms and compounds analysis and question split based on length.
\citet{Finegan-Dollak2018} proposes a query template-based split with word substitution that was much more challenging than the question split.
However, these splits are limited by the dataset content, making it difficult to construct a challenging benchmark while ensuring that every question phrase (sub-sentence) appears in the training set.

\begin{figure}[t]
  \includegraphics[width=0.47\textwidth]{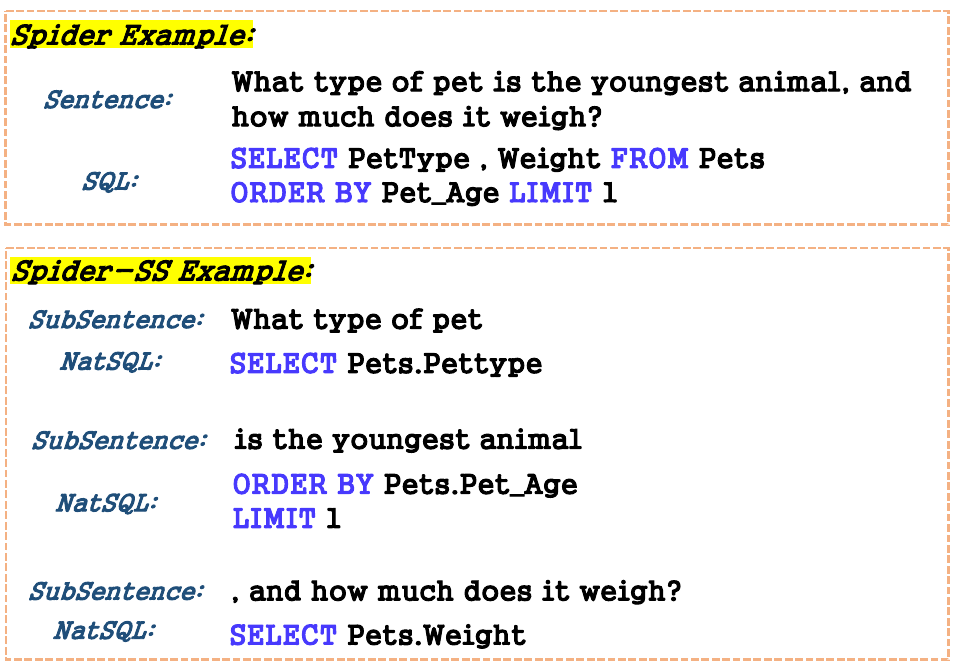}
  \centering
  \caption{A natural language sentence in the original Spider benchmark is split into three sub-sentences in Spider-SS, where each sub-sentence has a corresponding NatSQL clause.}
  \label{figure:Spider-SS}
\end{figure}

Previous works~\cite{NEURIPS2020_12b1e42d,wang2021structured,NEURIPS2020_83adc922} improve generalization by enhancing the model's component awareness.
Similarly, \citet{yin-etal-2021-compositional} and \citet{herzig-berant-2021-span} propose span-based semantic parsers that predict a sub-program over an  utterance span.
However, these works are based on datasets where component alignment is relatively easy to achieve; but
for more complex text-to-SQL, their methods cannot be used directly.
For example, as shown in the lower part of Figure \ref{figure:Spider-SS}, to align the sub-sentence with the sub-SQL, the algorithm needs to know that `\emph{youngest}' corresponds to `\emph{age}', and `\emph{weigh}' corresponds to `\emph{weigh\underline{t}}'.
For small or single-domain settings, such an alignment algorithm can be built by establishing rules; 
however, there is currently no simple and feasible alignment method for large complex cross-domain text-to-SQL, as in e.g.\ the Spider benchmark~\cite{Yu2018a}.


In this work, we first introduce a new dataset, Spider-SS (SS stands for \textit{sub-sentence}), derived from Spider \cite{Yu2018a}; Figure~\ref{figure:Spider-SS} compares the two.
To build Spider-SS, we first design a sentence split algorithm to split every Spider sentence into several sub-sentences until indivisible.
Next, we annotate every sub-sentence with its corresponding SQL clause,
reducing the difficulty of this task by using the intermediate representation language NatSQL \cite{gan2021natural}, which is simpler and syntactically aligns better with natural language (NL).
Spider-SS thus provides a new resource for designing models with better generalization capabilities without designing a complex alignment algorithm.
Furthermore, it can also be used as a benchmark for evaluating future alignment algorithms.
To our knowledge, this is the first sub-sentence-based text-to-SQL dataset.

\begin{figure}[t]
  \includegraphics[width=0.47\textwidth]{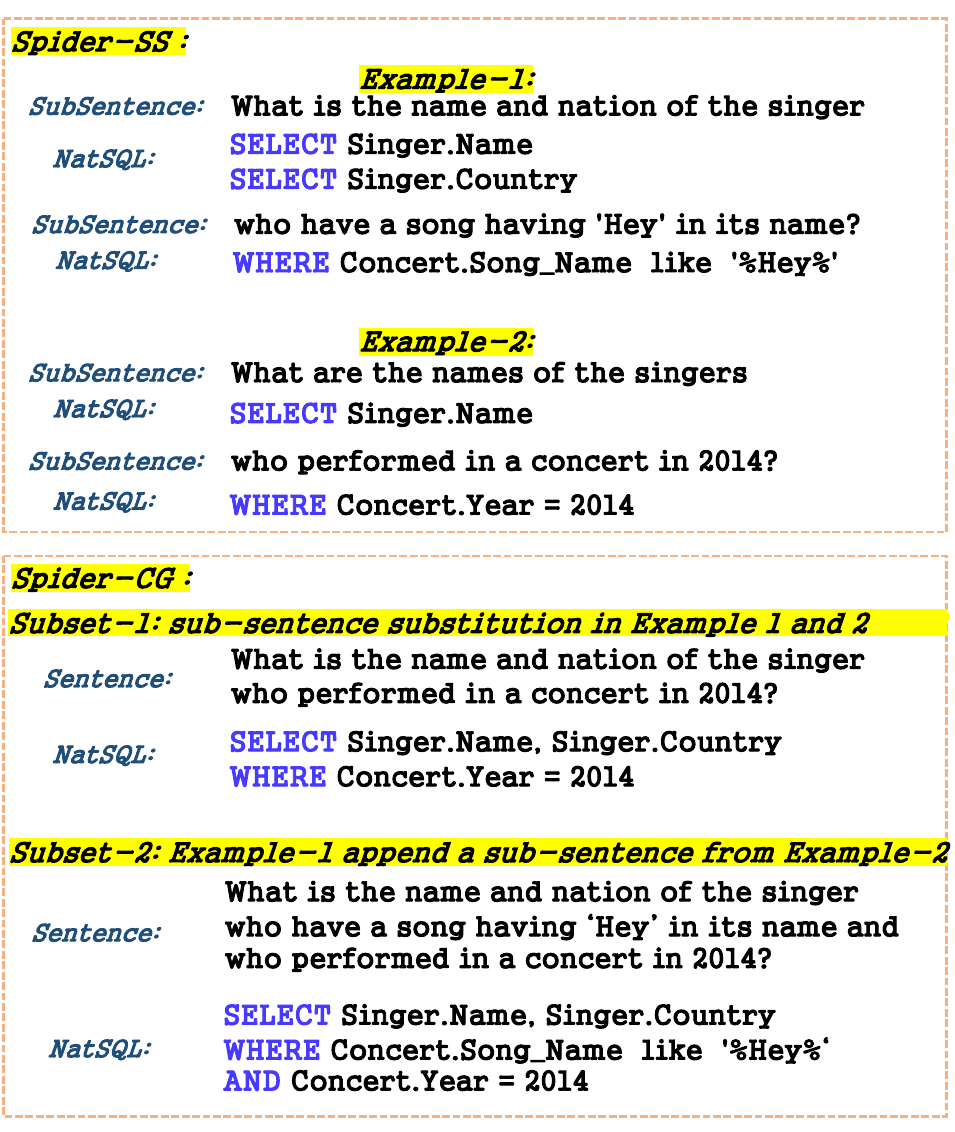}
  \centering
  \caption{Two Spider-CG samples generated by: (1) substituting the sub-sentence with one from another example; or (2) composing sub-sentences from 2 examples in Spider-SS.}
  \label{figure:Spider-CG}
\end{figure}

Our annotated Spider-SS provides us with sub-sentences paired with NatSQL clauses, which serve as our elements. Based on Spider-SS, we then construct a further dataset Spider-CG (CG stands for \textit{compositional generalization}), by substituting sub-sentences with those from other samples, or composing two sub-sentences to form a more complicated sample.
Spider-CG contains two subsets; Figure~\ref{figure:Spider-CG} shows one example for each.
The first subset contains 23,569 examples generated by substituting sub-sentences; we consider most data in this subset as in-distribution. 
The second subset contains 22,030 examples generated by appending sub-sentences,  increasing the length and complexity of the sentence and the SQL query compared to the original samples; we consider this subset as OOD.
We demonstrate that when models are trained only on the original Spider dataset, they suffer a significant performance drop on the second OOD subset of Spider-CG, even though the domain appears in the training set.

To improve the generalization performance of text-to-SQL models, we modify several previous state-of-the-art models so that they can be applied to the Spider-SS dataset, with the model trained sub-sentence by sub-sentence.
This modification obtains more than 7.8\%  accuracy improvement on the OOD subset of Spider-CG.

In short, we make the following contributions: 

\begin{itemize}[itemsep=0em,topsep=0em]
    \item 
    Besides the sentence split algorithm, we introduce Spider-SS, a human-curated sub-sentence-based text-to-SQL dataset built upon the Spider benchmark, by splitting its NL questions into sub-sentences. 
    \item We introduce the Spider-CG benchmark for measuring the compositional generalization performance of text-to-SQL models.  
    \item 
    We show that text-to-SQL models can be adapted to sub-sentence-based training, improving their generalization performance.
\end{itemize}

\section{Spider-SS}
\subsection{Overview}
Figure \ref{figure:Spider-SS} presents a comparison between Spider and Spider-SS.
Unlike Spider, which annotates a whole SQL query to an entire sentence, Spider-SS annotates the SQL clauses to sub-sentences.
Spider-SS uses NatSQL~\cite{gan2021natural} instead of SQL for annotation, because it is sometimes difficult to annotate the sub-sentences with corresponding SQL clauses due to the SQL language design.
The Spider-SS provides a combination algorithm that collects all NatSQL clauses and then generates the NatSQL query, where the NatSQL query can be converted into an SQL query.

The purpose of building Spider-SS is to attain clause-level text-to-SQL data avoiding the need for an alignment algorithm that is hard to build based on the complex large cross-domain text-to-SQL dataset, e.g., Spider benchmark. 
Besides, we can generate more complex examples through different combination of clauses from Spider-SS.
Consistent with Spider, Spider-SS contains 7000 training and 1034 development examples, but Spider-SS does not contain a test set since the Spider test set is not public. 
There are two steps to build Spider-SS. 
First, design a sentence split algorithm to cut the sentence into sub-sentences, and then manually annotate the NatSQL clause corresponding to each sub-sentence.

\subsection{Sentence Split Algorithm}
\label{section:Sentence Split Algorithm}

\begin{figure}[t]
  \includegraphics[width=0.47\textwidth]{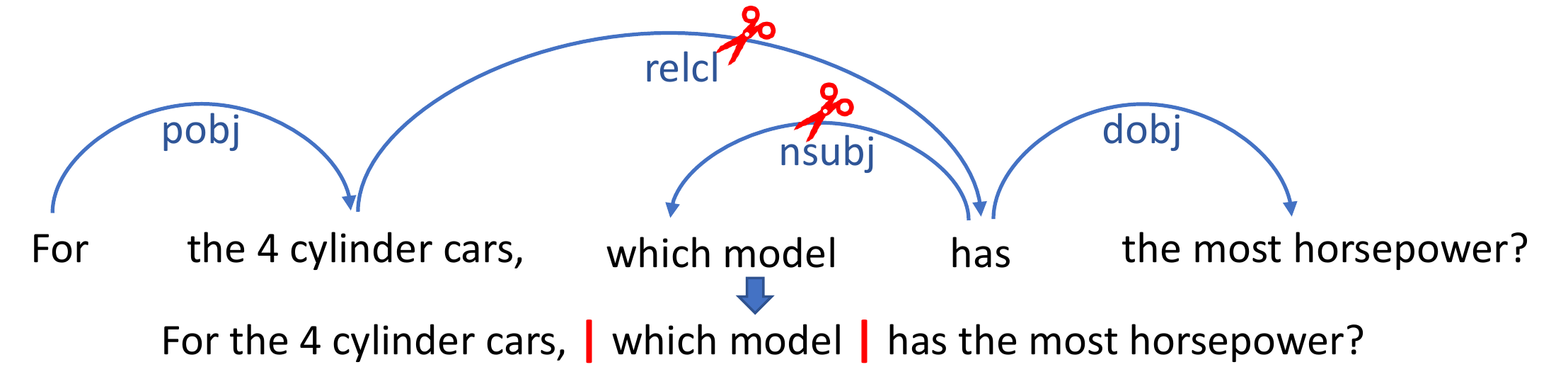}
  \centering
  \caption{Dependency structure of a sentence and how to split this sentence into three sub-sentences.}
  \label{figure:split algorithm}
\end{figure}
\begin{figure}[t]
  \includegraphics[width=0.47\textwidth]{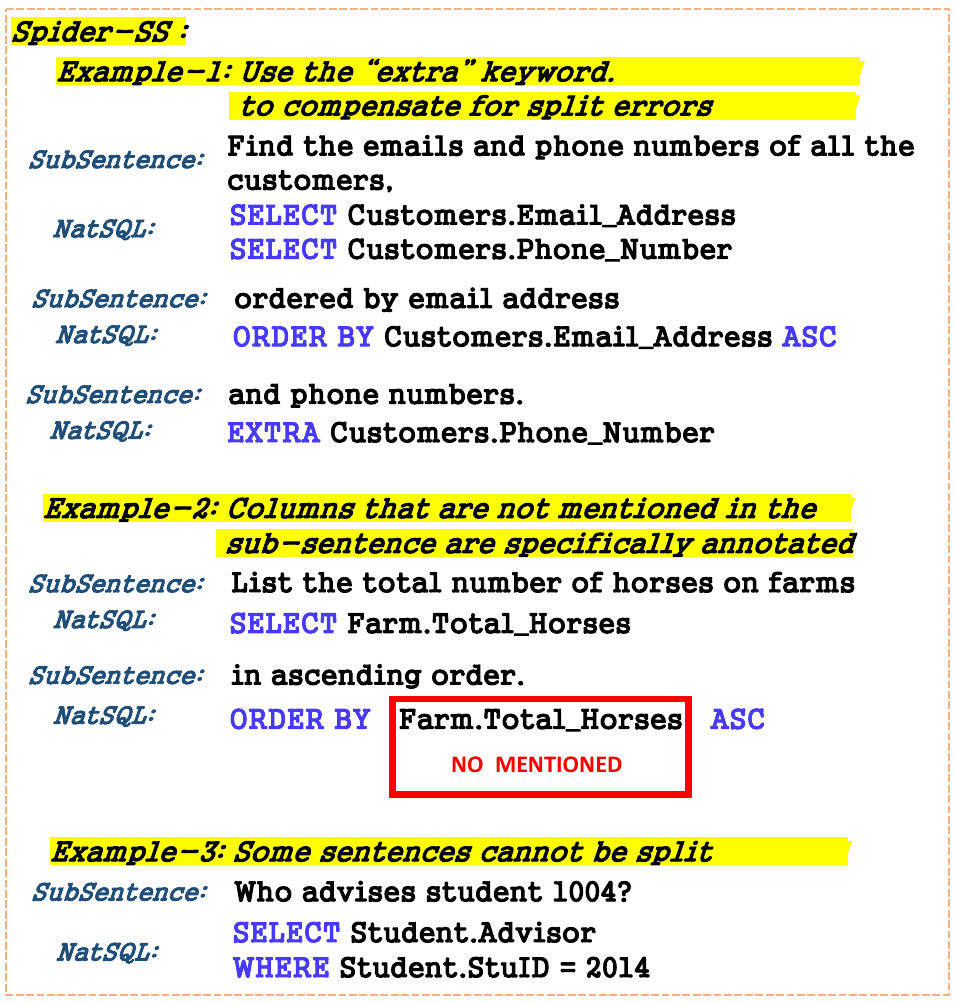}
  \centering
  \caption{Spider-SS examples in three special cases.}
  \label{figure:spider-ss-3expample}
\end{figure}

We build our sentence split algorithm upon the NL dependency parser spaCy~\footnote{https://github.com/explosion/spaCy}, which provides the grammatical structure of a sentence.  
Basically, we split the sentence with the following dependencies: \emph{
prep, relcl, advcl, acl, nsubj, npadvmod, csubj, nsubjpass} and \emph{conj}.
According to \cite{Marnee2010StanfordTD}, these dependencies help us separate the main clause, subordinate clauses, and modifiers.
Figure \ref{figure:split algorithm} shows the dependency structure of a sentence and how to split this sentence into three sub-sentences.
However, not every sentence would be split since there are some non-splittable sentences, such as the third example in Figure \ref{figure:spider-ss-3expample}, with the same annotation as the Spider dataset.
Although this method can separate sentences well in most cases, due to the variability of natural language, some examples cannot be perfectly split. 

To address the remaining issues in sentence split, we design some refinement steps tailored to text-to-SQL applications. For example, when the phase of a schema column or table is accidentally divided into two sub-sentences, these two sub-sentences are automatically concatenated.
Besides, when there is only one word in a sub-sentence, the corresponding split should also be undone.

We sampled 500 examples from the Spider-SS development set to evaluate the acceptability of splitting results manually, and only $<3\%$ of the splitting results are unsatisfactory.
For example, in the splitting results of the first example in Figure \ref{figure:spider-ss-3expample}, the last two sub-sentence should be combined to correspond to ``{\bf ORDER BY } Customer.Email\_Address, Customer.Phone\_Number {\bf ASC }''.
In this example, we did not simply give an ``{\bf ORDER BY } Customer.Phone\_Number {\bf ASC }'' to the last sub-sentence, because it does not mention anything related to ``{\bf ORDER BY }''.
Here, we introduce ``\emph{extra}'', a new NatSQL keyword designed for the Spider-SS dataset, indicating that this sub-sentence mentions a column that temporarily does not fit in any other NatSQL clauses.
When combining NatSQL clauses into the final NatSQL query, the combining algorithm determines the final position for the ``\emph{extra}'' column based on the clauses before and after. Note that even if there is a small proportion of unsatisfactory splitting results, as long as the model trained on Spider-SS can give the correct output according to the input sub-sentence, the quality of the sub-sentences itself does not strongly affect the model utility.

\subsection{Data Annotation}
\label{Section:data-annotation}
When we get the split results from the last step, we can start data annotation.
We give precise annotations based on the sub-sentence content, such as the ``\emph{extra}'' column annotation discussed in the last subsection.
Besides, if the description of the schema column is missing in the sub-sentence, we will give the schema column an additional  ``\emph{NO MENTIONED}''  mark.
For example, in the second example of Figure \ref{figure:spider-ss-3expample}, the ``\emph{in ascending order}'' sub-sentence does not mention the ``\emph{Farm.Total\_Horses}'' column. 
Therefore, we add a ``\emph{NO MENTIONED}'' mark for it.
For those sub-sentences that do not mention anything related to the query, we give a ``\emph{NONE}'' mark, representing there are no NatSQL clauses.

Since the annotation is carried out according to the sub-sentence content, the equivalent SQL that is more consistent with the sub-sentence will be preferred to the original SQL.
Similarly, if the original SQL annotation is wrong, we correct it according to the content.

\begin{figure}[t]
  \includegraphics[width=0.47\textwidth]{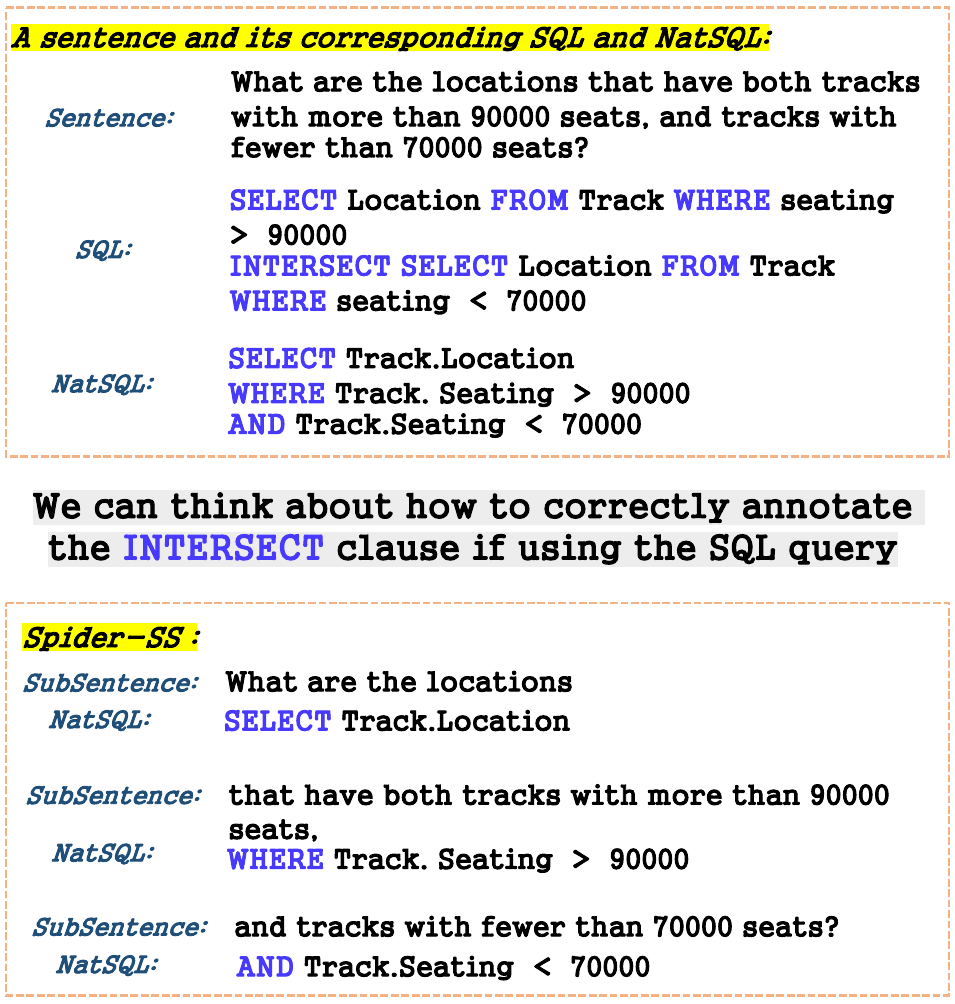}
  \centering
  \caption{It is difficult to annotate if using the SQL instead of NatSQL.}
  \label{figure:spider-ss+NatSQL}
\end{figure}

We annotate the sub-sentence using NatSQL instead of SQL, where NatSQL is an intermediate representation of SQL, only keeping the \textit{SELECT, WHERE, and ORDER BY} clauses from SQL.
Since some sub-sentences need to be annotated with \textit{GROUP BY} clause, we choose the version of NatSQL augmented with \textit{GROUP BY}.
We did not use SQL directly because it is difficult to annotate in some cases, such as the SQL example in Figure~\ref{figure:spider-ss+NatSQL}.
The difficulty is that there are two \textit{SELECT} clauses in this SQL query, but none of the sub-sentences seem to correspond to two \textit{SELECT} clauses.
In addition, considering that the two \textit{WHERE} conditions correspond to different \textit{SELECT} clauses, the annotation work based on SQL is far more difficult to complete.
As shown in Figure~\ref{figure:spider-ss+NatSQL}, we can use NatSQL to complete the annotation quickly, while the NatSQL can be converted back to the target SQL.
The detail of the annotation steps can be found in Appendix~\ref{sec:appendix-SS-Annotation}.

\section{Spider-CG}
\subsection{Overview}
\label{section:cg-Overview}
Spider-CG is a synthetic dataset, which is generated by recombining the sub-sentences of Spider-SS.
There are two recombination methods. The first is sub-sentence substitution between different examples, and the other is to append a sub-sentence into another sentence.
To facilitate the follow-up discussion, we named the Spider-CG subset generated by the sub-sentence substitution method {\bf CG-SUB}, and the other named {\bf CG-APP}.

In CG-SUB, there are 20,686 examples generated from the Spider-SS training set, while 2,883 examples are generated from the development set.
In  CG-APP, examples generated from training and development sets are 18,793 and 3,237, respectively.
Therefore, the Spider-CG contains 45,599 examples, around six times the Spider dataset.
We can further append sub-sentences to the CG-SUB examples if more data is needed.

\begin{algorithm*}[t]
  \caption{Generate CG-SUB and CG-APP dataset in a certain domain}\label{alg:infer}
  \hspace*{0.02in} {\bf Input:} 
  $e\_list$ \Comment{All compositional elements in a domain}
  \\
  \hspace*{0.02in} {\bf Output:} 
  $cg\_sub$ and $cg\_app$ \Comment{CG-SUB and CG-APP dataset in a certain domain}
  \begin{algorithmic}[1]
  \For {Every $element_1$ in $e\_list$}
  \For {Every $element_2$ in $e\_list$}
    \If {$element_1$ != $element_2$ }
\If {$element_1$.can\_be\_substituted\_by( $element_2$ ) }
    \State $cg\_sub$.append( $element_1$.generate\_substitution\_example( $element_2$ ) )
  \EndIf  
  
  \If {$element_1$.can\_append( $element_2$ ) }
    \State $cg\_app$.append( $element_1$.generate\_appending\_example( $element_2$ ) )
  \EndIf  
  
  \EndIf
  \EndFor
  \EndFor

  \State \textbf{return} $cg\_sub$, $cg\_app$ 
  \end{algorithmic}
  \end{algorithm*}

\subsection{Generation Algorithm}
According to Algorithm~\ref{alg:infer}, we can generate the CG-SUB and CG-APP based on compositional elements.
Each element contains one or more sub-sentences with corresponding NatSQL clauses from Spider-SS, where these NatSQL can only be \textit{WHERE or ORDER BY} clauses.
Thus, Algorithm~\ref{alg:infer} only substitute and append the \textit{WHERE and ORDER BY} clauses, and does not modify the \textit{SELECT} clause.
We collect the sub-sentences for compositional elements by scanning all sub-sentence from start to end or from end to start and stopping when encountering clauses except \textit{WHERE and ORDER BY}.
For example, we generate a compositional element containing the last two sub-sentences of the Spider-SS example in Figure~\ref{figure:spider-ss+NatSQL}.
In contrast, no element is extracted from the example in Figure~\ref{figure:Spider-SS}.
It should be noted that elements in a domain cannot be used in another because the schema items are different.
So as many domains as there are, it needs to run  Algorithm~\ref{alg:infer} as many times.
We recommend reading Appendix~\ref{sec:appendix-A} for details of \emph{can\_be\_substituted\_by} and \emph{can\_append} functions.

\begin{table}[t]
  \centering
  \resizebox{.99\columnwidth}{!}{
  \smallskip\begin{tabular}{cl}
    Ques & Show the name of employees  
    \\ & named Mark Young ?\\
    \hline
    SQL & {\bf SELECT} name {\bf FROM} employee  \\ &{\bf WHERE} name = `Mark Young'\\
    \hline
  \end{tabular}
  }
  \caption{One acceptable but not perfect examples in the Spider-CG.}
  \label{table:bad example of spider-ss}
  \end{table}

\subsection{Quality Evaluation}
\label{sec:Quality Evaluation}
We consider that the quality of a text-to-SQL sentence is determined by two criteria: containing the required information and being reasonable.
The `information' criterion requires a sentence that contains all the information needed to derive the target NatSQL.
The `reasonable' criterion requires a sentence that is logically correct and whose representation is fluent and easy to understand.
We randomly sampled 2000 examples from the Spider-CG dataset, around 99\% of which are acceptable, i.e., they meet the two criteria.
The evaluation is conducted manually by a computer science graduate with good knowledge of text-to-SQL.
However, these acceptable examples do not mean that there are no grammatical errors and they may be meaningless.
We give one acceptable but not perfect examples in Table \ref{table:bad example of spider-ss}, where the sentence is meaningless because the content it wants to query is the condition it gave. 
Besides, there are around 5\% NatSQL queries in these acceptable examples that can not be converted to the correct SQL.
This problem can be solved by a well-designed database schema or updating the NatSQL conversion function in the future.

\begin{figure}[t]
  \includegraphics[width=0.47\textwidth]{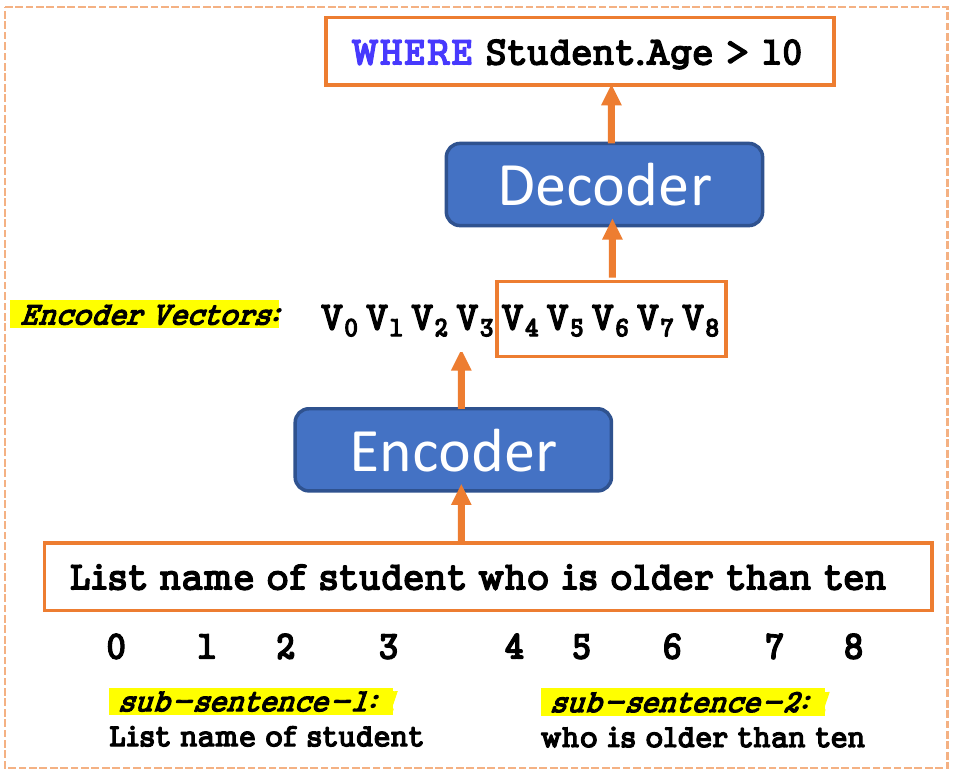}
  \centering
  \caption{A example of encoding the whole sentence but decoding only the sub-sentence.}
  \label{figure:model-work-flow}
\end{figure}

\section{Model}
\label{Section:model}
Existing text-to-SQL models input a sentence and output the corresponding SQL query.
So the easiest way to think of using the Spider-SS dataset is to train the model where inputting sub-sentence and outputting the corresponding NatSQL clauses.
However, this method is not workable because it will lose some essential schema information.
For example, if you only look at the third sub-sentence in Figure \ref{figure:Spider-SS}, you do not know whether it enquires about the weight of pets or people.

In order to take into account the context and the sub-sentence data of Spider-SS, we propose that a seq2seq model can encode the whole sentence but decode only the sub-sentence.
Figure \ref{figure:model-work-flow} presents the workflow of encoding the whole sentence but only decoding the sub-sentence of `\emph{who is older than ten}' and outputting the corresponding NatSQL clause.
Based on this modification, a seq2seq text-to-SQL model can be adapted to the Spider-SS.
Although previous span-based semantic parsers~\cite{yin-etal-2021-compositional,herzig-berant-2021-span} can work with aligned annotations based on the Spider-SS dataset, none of them are designed for complex text-to-SQL problems.
Our modification idea is similar in principle to the span-based semantic parsers, but we did not change the existing model according to the span-based because our modification idea has a smaller workload.

\begin{figure}[t]
  \includegraphics[width=0.47\textwidth]{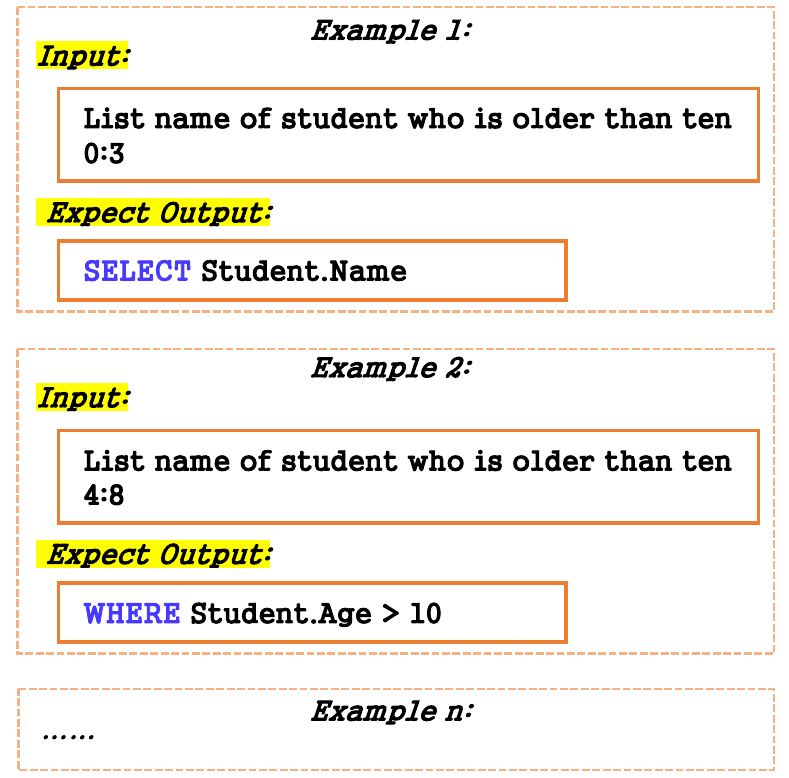}
  \centering
  \caption{A Spider-SS example is split into two examples for training and evaluation.}
  \label{figure:spider-ss-split-examples}
\end{figure}

In general, we can make the seq2seq-based text-to-SQL models adapt to the Spider-SS in three steps.
(1) Data preprocess. Split the Spider-SS examples by sub-sentence. For example, the example in Figure~\ref{figure:model-work-flow} is split to two examples shown in Figure~\ref{figure:spider-ss-split-examples}. 
(2) Model modification. After data preprocessing, there are two input data for a model. The first input is an entire question that directly goes to the encoder. The second input is the sub-sentence indexes, which are used to select the encoder output, as shown in Figure~\ref{figure:model-work-flow}.
(3) Output combination. Since the model output may be only a clause, not a complete NatSQL query, we generate the final NatSQL query after the model outputting all NatSQL clauses.

\section{Experiment}
\subsection{Experimental Setup}



\paragraph{Dataset.}
We evaluate the previous state-of-the-art models on the Spider-CG and Spider \cite{Yu2018a} datasets.
Since the Spider test set is not publicly accessible, Spider-CG does not contain a test set.
As discussed in Section \ref{section:cg-Overview}, we divide the Spider-CG into two subsets: CG-SUB and CG-APP.
Therefore, there are five evaluation sets: 
\begin{itemize}[leftmargin=*,noitemsep,topsep=0em]
    \item $\textbf{Spider}_{\textnormal{\small{D}}}$: the original Spider development set with 1,034 examples for \emph{cross-domain in-distribution} text-to-SQL evaluation.
    
    \item $\textbf{CG-SUB}_{\textnormal{\small{T}}}$:  the CG-SUB training set, containing 20,686 examples generated from Spider-SS training set by substituting sub-sentences. $\textnormal{CG-SUB}_{\textnormal{\small{T}}}$ can be used for \emph{in-domain in-distribution} text-to-SQL evaluation.
    
    \item $ \textbf{CG-SUB}_{\textnormal{\small{D}}}$:  the CG-SUB development set containing 2,883 examples for \emph{cross-domain in-distribution} text-to-SQL evaluation.
    
    \item $\textbf{CG-APP}_{\textnormal{\small{T}}}$:  the CG-APP training set, containing 18,793 examples generated from Spider-SS training set by appending sub-sentences. $\textnormal{CG-APP}_{\textnormal{\small{T}}}$ can be used for \emph{in-domain out-of-distribution}~\footnote{Out-of-distribution means that the difficulty distribution is different from the Spider; see Table~\ref{table:hardness}. Appendix \ref{sec:appendix-A} discusses the removal of overly complex examples to ensure that Spider-CG's SQL does not exceed the complexity upper bound of the Spider. } text-to-SQL evaluation.
    
    \item $ \textbf{CG-APP}_{\textnormal{\small{D}}}$: the CG-APP development set containing 3,237 examples for \emph{cross-domain out-of-distribution} text-to-SQL evaluation.
\end{itemize}

Our evaluation is based on the exact match 
metric defined in the original Spider benchmark.
The exact match metric measures whether the syntax tree of the predicted query without condition values is the same as that of the gold query. 
All models are only trained on 7000 Spider or Spider-SS training examples.

\paragraph{Models.}
We evaluate the following open-source models that reach competitive performance on Spider: 
\begin{itemize}[leftmargin=*,noitemsep,topsep=0em]
    \item $ \textbf{GNN}$:  The GNN~\cite{Bogin2019} model using the GLOVE~\cite{pennington-etal-2014-glove} embeddings.
    \item $ \textbf{RATSQL}$: The RATSQL~\cite{Wang2019} model using the GLOVE embeddings.
    \item $ \textbf{RATSQL}_{\textnormal{\small{B}}}$:  The RATSQL model using the BERT~\cite{Kenton2017} embeddings.
    \item $ \textbf{RATSQL}_{\textnormal{\small{G}}}$:  The RATSQL model using the GAP~\cite{DBLP:journals/corr/abs-2012-10309} embeddings.

    \item $_\textbf{(N)}$:  This subscript indicates that the model use NatSQL instead of SQL.
    \item $_\textbf{(S)}$:  This subscript indicates that the model is modified according to Section \ref{Section:model} and trained on Spider-SS. Besides, since Spider-SS is annotated by NatSQL, this subscript also indicates that the model uses NatSQL instead of SQL. 
\end{itemize}


\paragraph{Implementations.}
All experiments were performed on a machine with an Intel i5 9600 3.1GHz processor and a 24GB RTX3090 GPU.
All models keep their original hyperparameters except the
$\textnormal{RATSQL}_{\textnormal{\small{B(S)}}}$.
$\textnormal{RATSQL}_{\textnormal{\small{B(S)}}}$ cannot converge on the original parameters until we reduce the learning rate of model from 7.444e-04 to 1e-04 and raise the learning rate of BERT from 3e-06 to 1e-05.
We did not conduct a hyperparameter search, so the model trained on Spider-SS may improve performance through other parameters. 

\begin{table}[t]
    \centering
    \resizebox{.99\columnwidth}{!}{
    \smallskip\begin{tabular}{lcc}
        \hline
       \bf Dataset \ \ \ \ & \bf  Exact Match  & \bf  Execution Match  \\
        \hline \hline
         Training Set  & 90.7\%&  93.3\%     \\ 
         Development Set &  94.8\% &  95.2\%   \\
        \hline

    \end{tabular}
    }
    \caption{Use exact match and execution match metrics to evaluate the difference between the SQL in Spider and the SQL generated by NatSQL in Spider-SS. }\smallskip
    \label{table:gold-eval}
\end{table}

\subsection{Dataset Analysis}
\paragraph{Spider-SS.} 
\label{Section:data-analysis}

Table \ref{table:gold-eval} presents the difference between the SQL in Spider and the SQL generated by NatSQL in Spider-SS.
Our evaluation results are lower than the original NatSQL dataset \cite{gan2021natural} because the Spider-SS uses equivalent SQL and corrects some errors, as discussed in Section \ref{Section:data-annotation}.
Some equivalent and corrected SQL cannot get positive results in exact match metric and execution match.
Therefore, the model trained on Spider-SS may not be ideal for chasing the Spider benchmark, especially based on the exact match metric.
Similarly, the $ \textnormal{RATSQL}_{\textnormal{\small{G}}}$ extending NatSQL had achieved a previous SOTA result in the execution match of the Spider test set but get a worse result than the original in the exact match \cite{gan2021natural}.
Thus, we recommend using NatSQL-based datasets to evaluate models trained on NatSQL.

\begin{table}[t]
    \centering
    \resizebox{.99\columnwidth}{!}{
    \smallskip\begin{tabular}{lcccc}
        \hline
       \bf Dataset & \bf easy  & \bf  medium &\bf  hard  & \bf extra \\
        \hline \hline
        $\textbf{Spider}_{\textnormal{\small{D}}}$  & 24.1\% & 43.1\% & 16.8\% & 16.1\%\\ 
         $\textbf{CG-SUB}_{\textnormal{\small{T}}}$  & 28.6\% &  38.0\%  & 21.1\% &  12.3\% \\ 
         $\textbf{CG-SUB}_{\textnormal{\small{D}}}$ &  37.6\% &  38.4\% &   12.0\% &  12.0\%  \\
         $\textbf{CG-APP}_{\textnormal{\small{T}}}$ &  3.3\% & 31.4\% &  26.0\% &  39.3\%  \\ 
       $\textbf{CG-APP}_{\textnormal{\small{D}}}$ &  2.4\% &   44.3\% &  22.9\% & 30.4\% \\ 
        \hline

    \end{tabular}
    }
    \caption{The difficulty distribution of five different evaluation sets.}\smallskip
    \label{table:hardness}
\end{table}

\paragraph{Spider-CG.} 
Table \ref{table:hardness} presents the difficulty distribution of five different evaluation sets.
The difficulty criteria are defined by Spider benchmark, including \emph{easy, medium, hard} and \emph{extra hard}.
Experiments show that the more difficult the SQL is, the more difficult it is to predict correctly \cite{Wang2019,DBLP:journals/corr/abs-2012-10309,gan2021natural}.
It can be found from Table~\ref{table:hardness} that the difficulty distribution of $\textnormal{CG-SUB}_{\textnormal{\small{T}}}$ and $\textnormal{CG-SUB}_{\textnormal{\small{D}}}$ is similar to that of $\textnormal{Spider}_{\textnormal{\small{D}}}$.
The similar distributions among $\textnormal{CG-SUB}_{\textnormal{\small{T}}}$, $\textnormal{CG-SUB}_{\textnormal{\small{D}}}$, and $\textnormal{Spider}_{\textnormal{\small{D}}}$ support the view discussed in Section \ref{Section:intro} that the examples generated by the substitution method are in-distribution.

On the other hand, the difficulty distributions of $\textnormal{CG-APP}_{\textnormal{\small{T}}}$ and $\textnormal{CG-APP}_{\textnormal{\small{D}}}$ are obviously different from that of $\textnormal{Spider}_{\textnormal{\small{D}}}$.
Due to appending the sub-sentence, the NL and SQL in CG-APP become more complex, where the proportion of SQL in \emph{extra hard} increased significantly, while \emph{easy} was the opposite.

\subsection{Sentence Split Algorithm Evaluation}
We generate the Spider-CG based on the combination of Spider-SS sub-sentences split by the algorithm introduced in Section \ref{section:Sentence Split Algorithm}.
We can reuse this algorithm to split the sentence in Spider-CG and then compare the splitting results with the Spider-SS sub-sentences to evaluate the stability of the splitting algorithm.
We consider that a deviation of one or two tokens in the splitting result is acceptable.
For example, in Figure \ref{figure:Spider-SS}, we consider that putting the comma of the third sub-sentence into the second sub-sentence does not change the meaning of sub-sentences, same for moving both the comma and the word `and'.

Table \ref{table:split-eval} presents the similarity between sub-sentences in Spider-SS and Spider-CG, which are generated by the same split algorithm under the deviation of one or two words.
The similarity exceeds 90\% in all evaluation set when two deviation words are allowed.
Considering that the model trained on the Spider-SS does not require consistent split results, as discussed in Section \ref{section:Sentence Split Algorithm},  the similarity results of the splitting algorithm are good enough.
The similarity of CG-SUB is higher than that of CG-APP, which means the more complex the sentence, the greater the challenge to the algorithm.
Although the algorithm has been refined on the training set, the similarity between training and development in CG-SUB and CG-APP is close, showing that the algorithm performs consistently for sentences in unseen domains.
In summary, we consider that as long as the sentences are not more complex than CG-APP, the algorithm can be used stably in other text-to-SQL datasets.

\begin{table}[t]
    \centering
    \resizebox{.99\columnwidth}{!}{
    \smallskip\begin{tabular}{lcc}
        \hline
       \bf Dataset \ \ \ \ & \bf  Deviation <= 1  & \bf  Deviation <= 2 \\
        \hline \hline
         $\textbf{CG-SUB}_{\textnormal{\small{T}}}$  & 93.2\% &  94.4\%    \\ 
         $\textbf{CG-SUB}_{\textnormal{\small{D}}}$ &  92.9\% &  94.1\%   \\
         $\textbf{CG-APP}_{\textnormal{\small{T}}}$ &  86.0\% & 90.4\%  \\ 
       $\textbf{CG-APP}_{\textnormal{\small{D}}}$ &  88.9\% &   92.6\%  \\ 
        \hline

    \end{tabular}
    }
    \caption{The similarity between sub-sentences in Spider-SS and Spider-CG generated by the same split algorithm under the deviation of one or two tokens.  }\smallskip
    \label{table:split-eval}
\end{table}

\begin{table*}[t]
    \centering
    \resizebox{1.8\columnwidth}{!}{
    \smallskip\begin{tabular}{lccccc}
        \hline
       \bf Approach & \bf $\textbf{Spider}_{\textnormal{\small{D}}}$ & \bf $\textbf{CG-SUB}_{\textnormal{\small{T}}}$ & $\textbf{CG-SUB}_{\textnormal{\small{D}}}$ &  $\textbf{CG-APP}_{\textnormal{\small{T}}}$ &$\textbf{CG-APP}_{\textnormal{\small{D}}}$ \\
        \hline \hline
        
         $\textbf{RATSQL}_{\textnormal{\small{G}}}$  &  72.7\% &   80.9\% & 70.3\% & 45.2\% & 44.2\% \\ 
         $\textbf{RATSQL}_{\textnormal{\small{G(N)}}}$ &  73.9\% &  90.2\% &   75.0\% &  67.8\% &  60.5\% \\
         $\textbf{RATSQL}_{\textnormal{\small{G(S)}}}$ & \bf 74.5\% & \bf 91.4\% & \bf  76.7\% & \bf 82.5\% & \bf 68.3\% \\ 
       
        \hline
        $\textbf{RATSQL}_{\textnormal{\small{B}}}$ & 72.0\% & 79.5\% & 72.0\% & 45.1\% & 47.2\%\\ 
        $\textbf{RATSQL}_{\textnormal{\small{B(N)}}}$ & \bf 72.1\% &  83.2\% &   69.4\% &  54.6\% &  53.1\% \\
         $\textbf{RATSQL}_{\textnormal{\small{B(S)}}}$ &  71.9\% & \bf 91.0\% & \bf  72.6\% & \bf 79.8\% & \bf 61.5\% \\ 
        
        \hline
        
        $\textbf{RATSQL}_{\textnormal{\small{(N)}}}$ &  63.2\% &  79.1\% &   60.7\% &  40.6\% &  34.5\% \\
         $\textbf{RATSQL}_{\textnormal{\small{(S)}}}$ & \bf 64.7\% & \bf 88.8\% & \bf  63.3\% & \bf 72.1\% & \bf 44.1\% \\ 
        
        \hline
        
        $\textbf{GNN}_{\textnormal{\small{(N)}}}$ & \bf 54.4\% &  67.3\% &   \bf 57.5\% &  30.4\% &  25.1\% \\
         $\textbf{GNN}_{\textnormal{\small{(S)}}}$ &  49.3\% & \bf 71.9\% &   51.8\% & \bf 52.1\% & \bf 34.6\% \\ 
        
        \hline

    \end{tabular}
    }
    \caption{Exact match accuracy on evaluation sets.}\smallskip
    \label{table:Exact match}
\end{table*}

\subsection{Model Results}

Table \ref{table:Exact match} presents the exact match accuracy on the five different evaluation sets. 
In the two OOD datasets, $\textnormal{CG-APP}_{\textnormal{\small{T}}}$ and $\textnormal{CG-APP}_{\textnormal{\small{D}}}$, the performance of all models has dropped by about 10\% to 30\%.
However, the models trained on Spider-SS significantly outperform those trained on Spider when evaluated on the OOD datasets.
We use the sentence split algorithm to split every sentence before inputting the models with subscript (S). 
Although the split sub-sentences are not completely consistent with those seen during training, it did not prevent the models with subscript (S) from getting good performance, i.e.,
the $\textnormal{RATSQL}_{\textnormal{\small{G(S)}}}$ consistently outperforms all other models on all evaluation sets.
These results demonstrate that the sub-sentence-based method can improve the generalization performance.
The limitation is that the method may not be compatible with the original model, e.g., original hyperparameters in $\textnormal{RATSQL}_{\textnormal{\small{B(S)}}}$ are not workable, and the performance of GNN on the $\textnormal{Spider}_{\textnormal{\small{D}}}$ and $\textnormal{CG-SUB}_{\textnormal{\small{D}}}$ is degraded.

Each model has a close result between the unseen $\textnormal{Spider}_{\textnormal{\small{D}}}$ and $\textnormal{CG-SUB}_{\textnormal{\small{D}}}$, indicating that from the perspective of the model, the synthetic sentences are pretty similar to NL.
Therefore, we believe the performance on $\textnormal{CG-SUB}_{\textnormal{\small{D}}}$ can be generalized to the real world.
Moreover, considering that the algorithms for generating $\textnormal{CG-SUB}_{\textnormal{\small{D}}}$ and $\textnormal{CG-APP}_{\textnormal{\small{D}}}$ are close (see Appendix \ref{sec:appendix-A}), we can further speculate that the synthetic sentences of $\textnormal{CG-APP}_{\textnormal{\small{D}}}$ are also close to natural language. 



The models with NatSQL is significantly better than that without NatSQL when evaluated on Spider-CG.
One of the reasons is that the training data of Spider and Spider-SS are about 10\% different, which leads to the performance degradation in the model trained on Spider when evaluated on the SQL generated by the NatSQL of Spider-SS, and vice versa.
On the other hand, experiments in \cite{gan2021natural} show that NatSQL improve the model performance in \emph{extra hard} SQL.
Therefore, $\textnormal{RATSQL}_{\textnormal{\small{G(N)}}}$ and $\textnormal{RATSQL}_{\textnormal{\small{B(N)}}}$ suffer less performance degradation in $\textnormal{CG-APP}_{\textnormal{\small{T}}}$ and $\textnormal{CG-APP}_{\textnormal{\small{D}}}$ than $\textnormal{RATSQL}_{\textnormal{\small{G}}}$ and $\textnormal{RATSQL}_{\textnormal{\small{B}}}$.

\section{Limitation of this Work}
The Spider-SS and Spider-CG are based on Spider, an English large-scale text-to-SQL dataset, and we did not extend the experiment to other language and text-to-SQL datasets.
Therefore, we did not verify whether these methods work well in other languages and datasets.
Besides, since this work is based on NatSQL, there will be around 5\% of NatSQL that can not be converted to the correct SQL.
\section{Related Work}


\paragraph{Data augmentation for text-to-SQL models.} 
Data augmentation has been commonly used for improving performance~\cite{Xiong2019,Li2019}.
In the context of text-to-SQL generation, \citet{Yu2018-SyntaxSQLNet} generate synthetic training samples from some pre-defined SQL and NL question templates.
\citet{parikh-etal-2020-totto} introduces an table-to-text dataset with over 120,000 examples that proposes a controlled generation task: given a Wikipedia table and a set of highlighted table cells, produce a one-sentence description. 
\citet{yu2021grappa} sample from the given examples and then give a large number of tables to generate new synthetic examples.
\citet{DBLP:journals/corr/abs-2012-10309} present  a  model  pre-training framework that jointly learns  representations of NL utterances and table schemas by leveraging generation models to generate pre-train data.
Our proposed Spider-CG dataset can be used for data augmentation.

\paragraph{Compositional generalization for semantic parsing.} 
Compositional generalization for semantic parsing has captured wide attention recently \cite{Finegan-Dollak2018,oren-etal-2020-improving,DBLP:journals/corr/abs-2007-08970,conklin-etal-2021-meta}.
Most prior works on text-to-SQL tasks focus on the cross-domain generalization, which mainly assess how the models generalize the domain knowledge to new database schemas~\cite{Suhr2020,gan2021exploring}.
On the other hand, \citet{shaw-etal-2021-compositional} introduces TMCD splits for studying compositional generalization in semantic parsing, where they aim to maximize the divergence of SQL compounds between the training and test sets.

Although both the TMCD split and our Spider-CG can be used to evaluate the text-to-SQL compositional generalization ability, their problem setting is different.
TMCD split is based on SQL syntax structure, while Spider-CG is based on the natural language syntax, which leads to different requirements for compositional generalization ability.
For example, TMCD splits requires model learning ``\emph{Give me the name of students who is the oldest}'' can predict the ``\emph{Give me the name of the oldest student}'' since their SQL is the same.
Spider-CG does not expect the model to do so because the syntax of questions is different, i.e., ``\emph{Give me the name of students who is the oldest}'' contains two sub-sentences, and none of them is close to the ``\emph{Give me the name of the oldest student}''.
In other words, Spider-CG requires the model learning ``\emph{List the id of the oldest dog}'' can predict the ``\emph{Give me the name of the oldest student}''.

Our model is inspired by prior works on neural parsers constructed to capture granular information from a whole.
\citet{yin-etal-2021-compositional} describe a span-level supervised attention loss that improves compositional generalization in semantic parsers.
\citet{herzig-berant-2021-span} propose SpanBasedSP, a parser that predicts a span tree over an input utterance, and dramatically improves performance on splits that require compositional generalization.
\citet{NEURIPS2020_12b1e42d} propose the Neural-Symbolic Stack machine (NeSS), which integrates a symbolic stack machine into a seq2seq generation framework, and learns a neural network as the controller to operate the machine.
However, these works are based on datasets where component alignment is relatively easy to achieve; but for more complex text-to-SQL, their methods cannot be used directly.
Our proposed Spider-SS can be used to replace or evaluate the alignment algorithm.
\section{Conclusion}
We introduce Spider-SS and Spider-CG for measuring compositional generalization of text-to-SQL models.
Specifically, Spider-SS is a human-curated sub-sentence-based text-to-SQL dataset built upon the Spider benchmark.
Spider-CG is a synthetic text-to-SQL dataset constructed by substituting and appending sub-sentences of different samples, so that the training and test sets consist of different compositions of sub-sentences.
We found that the performance of previous text-to-SQL models drop dramatically on the Spider-CG OOD subset, while modifying the models to fit the segmented data of Spider-SS improves compositional generalization performance.

\section*{Acknowledgements}
We thank the anonymous reviewers for their helpful comments.
Matthew Purver acknowledges financial support from the UK EPSRC under grant EP/S033564/1, and from the Slovenian Research Agency for research core funding (No.\ P2-0103 and No.\ P5-0161). 
Xinyun Chen is supported by the Facebook Fellowship. 
\bibliography{custom}
\bibliographystyle{acl_natbib}

\appendix

\section{Further Discussion of Algorithm~\ref{alg:infer}}
\label{sec:appendix-A}

As discussed in Section \ref{sec:Quality Evaluation}, we need to ensure that the Spider-CG examples meet the criteria of containing required information and being reasonable.
To ensure that the generated Spider-CG sentence contains the required information, the compositional element needs to contain all the information needed to derive the target NatSQL clause. 
Thus some sub-sentence can not be a compositional element, such as the last sub-sentence of examples 1 and 2 in Figure~\ref{figure:spider-ss-3expample}. 
Among them, example 1 misses \textit{ORDER BY} information; example 2 misses \emph{Total\_Horses} column information.
In contrast, the sub-sentence of the two Spider-SS examples in Figure \ref{figure:Spider-CG} contains the required information and can be compositional elements.
So, we can filter out the sub-sentences containing the ``\emph{NO MENTIONED}'' and ``\emph{extra}'' label, and collect the rest as compositional elements.


The `\emph{can\_be\_substituted\_by}' and `\emph{can\_append}' function in Algorithm~\ref{alg:infer} are used to ensure that the generated sentences are reasonable.
For the convenience of discussion, we refer to them as `\emph{sub}' and `\emph{app}' functions for short. 
These two functions examine the generated sentences from complexity, logic and coherence.

\paragraph{Complexity} checks are used to limit the complexity of the generated examples to no more complex than the upper bound of the Spider dataset.
On the NatSQL side, both functions do not allow the generated NatSQL containing: 1) more than one subqueries; 2) more than one \textit{HAVING} condition; 3) more than three \textit{WHERE} conditions; 4) more than one \textit{ORDER BY} clause; 5) new conditions for a subquery.
On the NL side, since the substitution did not clearly increase the sentence complexity, only the `\emph{app}' function performs the NL complexity checks to restrict the number of sub-sentence to less than 4.

\paragraph{Logic} checks are used to prevent generating contradictory examples.
First, logic checks filter out examples with repeated \textit{WHERE} conditions. 
Then, it filters out examples whose \textit{WHERE} condition negates the query content, e.g., \emph{what is name of student that do not have any student}.
Finally, since the \textit{GROUP BY} clause is often expressed implicitly, substituting or appending elements containing the \textit{GROUP BY} clause may introduce logical errors. Thus, logic checks require the \textit{GROUP BY} clauses to be the same if they exist.


\paragraph{Coherence} checks are used to ensure that the expression of the generated sentence is coherent.
As discussed in Section \ref{section:Sentence Split Algorithm}, we separate a sentence into main clause, subordinate clauses, and modifiers. 
The main clause expresses what you want to query, i.e., corresponding to the SELECT clause. 
Subordinate clauses and modifiers are restrictions on the query, i.e., corresponding to \textit{WHERE} and \textit{ORDER BY} clauses. 
Therefore, compositional elements only contain subordinate clauses and modifiers.
The way to ensure the coherence of sentences by \emph{sub} function is to require the substitution sub-sentences modify the same noun.
Suppose the schema table of the NatSQL in a compositional element appears in advance. In that case, we consider its sub-sentence modifies the table noun because repeating a known object~\footnote{A table is usually an object whose attributes are its columns in relational databases.} can only be a further modification.
However, if the schema table has not appeared before, we consider that the sub-sentence modifies its previous word since a subordinate clause usually comes immediately after the noun it describes.

There is a high similarity between the \emph{app} and \emph{sub} function, but the inspection between the substituted elements is changed to the inspection between the new element and the last element in the original sentence. 
Therefore, the appended sub-sentence must modify the same noun as the last sub-sentence. 
If a compositional element passes the \emph{app} function, we use the word `\emph{and}' or `\emph{or}' to connect it where the word `\emph{or}' can only connect a \textit{WHERE} condition.
Table~\ref{table:cg-example-in-appendix} discuss some examples for ease of understanding.


\begin{table*}[t]
  \centering
  \resizebox{1.99\columnwidth}{!}{
  \smallskip\begin{tabular}{l}

    \\
     \hline
     Spider sentence:\\
     Show name for all singers ordered by age from the oldest to the youngest.\\
     How many concerts are there in year 2014 or 2015? \smallskip   \\
    
     Generate new sentence by appending: \\
     Show name for all singers ordered by age from the oldest to the youngest and in year 2014 or 2015? \smallskip  \\
     
     Coherence checks:\\
     Failed to pass the coherence checks due to the modified noun of the two sub-sentences being different. \\
     In the same way, the `\emph{Show name for all singers in year 2014 or 2015?}' can not pass.\\
    \hline
    
    Spider sentence:\\
    Show name for all singers ordered by age from the oldest to the youngest.\\
     What is the nation of the singer who have a song having ' Hey ' in its name? \smallskip\\
     Generate new sentence by appending: \\
     What is ... who have a song having ' Hey ' in its name and ordered by age from the oldest to the youngest. \smallskip \\
     Coherence checks:\\
     Pass the coherence checks. \\
     In the same way, the `\emph{what is ... singer ordered by age from the oldest to the youngest .}' also pass.\\
    \hline
    
    \hline
    Spider sentence:\\
     What are the titles of the books whose writer is not 'Elaine Lee'?\\
     List the writers who have written more than one book.  \smallskip \\
     Generate new sentence by appending: \\
     What are the titles of the books whose writer is not 'Elaine Lee' and who have written more than one book. \smallskip\\
     Coherence checks:\\
     Failed to pass the coherence checks due to the modified noun of the two sub-sentences being different. \\
     In the same way, the `\emph{What are the titles of the books who have written more than one book.?}' can not pass. \\
    \hline
    Spider sentence:\\
     List the writers who have written more than one book.\\
     Show writers who have published a book with price more than 40.  \smallskip \\
         Generate new sentence by appending and substituting:  \\
     List the writers who have written more than one book and who have published a book with price more than 40.\\
     List the writers who have written more than one book or who have published a book with price more than 40 .\\
     Show writers who have published a book with price more than 40 and who have written more than one book .\\
     Show writers who have published a book with price more than 40 or who have written more than one book.\\
     List the writers who have written more than one book.\\
     Show writers who have written more than one book. \smallskip \\
         Coherence checks:\\
     All these sentence pass the coherence checks. \\
    \hline
    
  \end{tabular}
  }
  \caption{Some examples of successful or unsuccessful passing the coherence checks.}
  \label{table:cg-example-in-appendix}
    \end{table*}

\section{Unseen SQL Structure Template in Spider-CG}
Although we limit the complexity of the generated examples lower than the upper bound of the Spider dataset, Spider-CG still contains unseen SQL structure templates.
For example, the NatSQL template `\textit{SELECT} COL \textit{WHERE} COL > VAL or count(TABLE.*) >=VAL \textit{GROUP BY} COL' and corresponding SQL can not be found in the original Spider.
The new templates may degrade the performance of models.

\section{Spider-SS Annotation Steps}
\label{sec:appendix-SS-Annotation}
We build an annotation tool to show the sub-sentence and sub-SQL split from a question-NatSQL pair. 
During annotation, the annotators select the corresponding sub-SQL for sub-sentences. 
In rare cases, if there is no suitable sub-SQL, the annotators would write a new one, such as the example-1 in Figure\ref{figure:spider-ss-3expample}.
We recruit two graduate students major in computer science to annotate the dataset manually.
They are trained with a detailed annotation guideline and some samples. 
One is allowed to start after his trial samples are approved by the whole team.
Each example is annotated twice. 
If the annotations are different, the final annotation will be decided by a discussion. 
If two annotators discuss and conclude that one of the annotations is wrong and the other is correct, the correct annotation is retained. 
Otherwise, the authors will annotate this example if no such conclusion can be drawn.

\section{Execution Match}

The execution match metric measures whether the query results from the predicted query are the same as the gold query results.
The original RATSQL can not generate the executable SQL until extending the NatSQL. 
The NatSQL2SQL conversion would analyze the utterance and generate executable SQL, irrelevant to the RATSQL model. 
Thus we only report the results of models with NatSQL.
Since the execution match is similar to the exact match,  we only report the top models in Table \ref{table:Execution match}.
Similar to the exact match, $\textnormal{RATSQL}_{\textnormal{\small{G(S)}}}$ outperform  other models in most evaluation set except on the CG-APP$_T$.

\begin{table}[t]
    \centering
    \resizebox{.99\columnwidth}{!}{
    \smallskip\begin{tabular}{lccccc}
        \hline
      \bf Approach & \bf $\textbf{Spider}_{\textnormal{\small{D}}}$ & \bf $\textbf{CG-SUB}_{\textnormal{\small{T}}}$ & $\textbf{CG-SUB}_{\textnormal{\small{D}}}$ &  $\textbf{CG-APP}_{\textnormal{\small{T}}}$ &$\textbf{CG-APP}_{\textnormal{\small{D}}}$ \\
        \hline \hline
        
         $\textbf{RATSQL}_{\textnormal{\small{G(N)}}}$ &  75.8\% & 86.7\% &   78.0\% & 70.4 \% &  68.9\% \\
$\textbf{RATSQL}_{\textnormal{\small{B(S)}}}$ &  74.7\% &  87.9\% &   76.4\% & \bf 82.0\% &  72.5\% \\ 
         $\textbf{RATSQL}_{\textnormal{\small{G(S)}}}$ & \bf 76.7\% & \bf 88.3\% & \bf  80.4\% &  78.8\% & \bf 75.1\% \\ 
       
        \hline

    \end{tabular}
    }
    \caption{Execution match accuracy on evaluation sets.}\smallskip
    \label{table:Execution match}
\end{table}

\end{document}